\documentclass{article}

\usepackage{arxiv}

\usepackage[utf8]{inputenc} 
\usepackage[T1]{fontenc}    
\usepackage{hyperref}       
\usepackage{url}            
\usepackage{booktabs}       
\usepackage{amsfonts}       
\usepackage{amsmath}        
\usepackage{nicefrac}       
\usepackage{microtype}      
\usepackage{lipsum}		
\usepackage{graphicx}
\usepackage{subfigure}
\usepackage[sort, numbers]{natbib}
\usepackage{doi}
\usepackage{gensymb}
\usepackage{balance}
\usepackage{enumitem}
\usepackage{titlesec}
\usepackage{multicol}
\usepackage{caption}
\usepackage{float}

\title{Face Recognition Using Synthetic Face Data}

\author{ {\hspace{1mm}Omer Granoviter} \\
	\texttt{omer.granoviter@datagen.tech} \\
	\And
	{\hspace{1mm}Alexey Gruzdev} \\
	\texttt{alexey.gruzdev@datagen.tech} \\
        \And
        {\hspace{1mm}Vladimir Loginov} \\
	\texttt{vladimir.loginov@datagen.tech} \\
         \And
	{\hspace{1mm}Max Kogan} \\
	\texttt{max.kogan@datagen.tech} \\
        \And
	{\hspace{1mm}Orly Zvitia} \\
	\texttt{orly.zvitia@datagen.tech} \\
}
\date{}

\renewcommand{\headeright}{}
\renewcommand{\undertitle}{}
\renewcommand{\shorttitle}{}

\hypersetup{
pdftitle={Synthetic Data},
pdfsubject={Synthetic Data},
pdfauthor={Omer Granoviter},
pdfkeywords={Synthetic Data, Datagen, Face Recognition},
}

\begin{document}

\maketitle

\begin{figure}[!ht]
    \centering
    \includegraphics[width=1\textwidth]{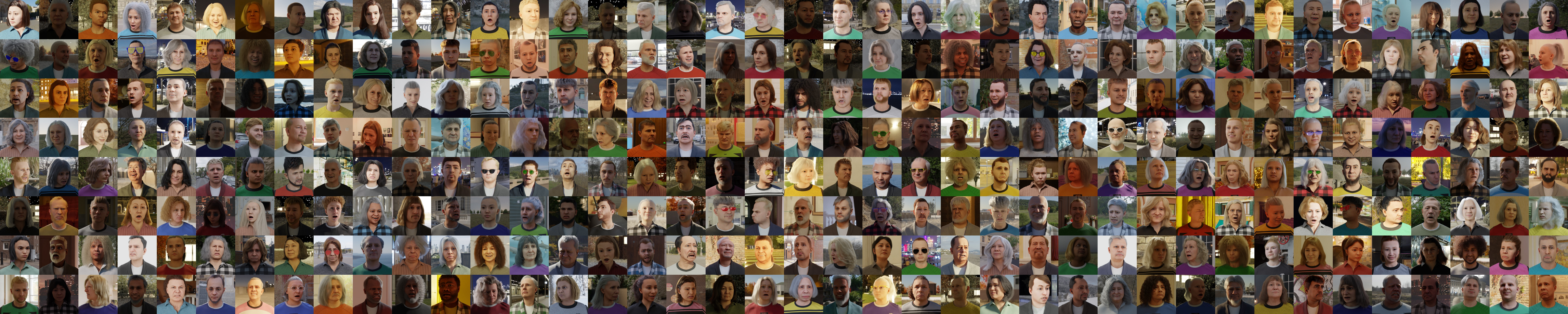}
    \caption{Example of randomly sampled 320 identities from the 30,000 used in the research. Our pipeline enables granular control of almost everything in the scene (e.g., pose, accessories, background, light, hair, eyebrows, eyes).}
    \label{fig:mosiac}
\end{figure}

\begin{abstract}
In the field of deep learning applied to face recognition, securing large-scale, high-quality datasets is vital for attaining precise and reliable results. 
However, amassing significant volumes of high-quality real data faces hurdles such as time limitations, financial burdens, and privacy issues. Furthermore, prevalent datasets are often impaired by racial biases and annotation inaccuracies. In this paper, we underscore the promising application of synthetic data, generated through rendering digital faces via our computer graphics pipeline, in achieving competitive results with the state-of-the-art on synthetic data across multiple benchmark datasets. By finetuning the model,
we obtain results that rival those achieved when training with hundreds of thousands of real images (98.7\% on LFW \cite{LFWTech}). We further investigate the contribution of adding intra-class variance factors (e.g., makeup, accessories, haircuts) on model performance. Finally, we reveal the sensitivity of pre-trained face recognition models to alternating specific parts of the face by leveraging the granular control capability in our platform.

\end{abstract}

\begin{multicols}{2}
\section{Introduction}
\label{section:introduction}
Modern face recognition architectures \cite{deng2019arcface, sphereface, adaface, an2022killing} have demonstrated exceptional performance on benchmark face recognition test sets such as Labeled Faces in the Wild (LFW) \cite{LFWTech} and Celebrities in Frontal-Profile in the Wild (CFP-FP) \cite{cfp-paper}. achieving accuracy as high as 99.85\%  and 99.5\%, respectively. Despite these impressive results, the main challenge for developing state-of-the-art (SOTA) industrial-ready applications does not necessarily lie in refining the algorithms but rather in obtaining relevant and large-scale datasets.

Publicly available datasets satisfy conditions such as pose variability, image quality conditions, lightning conditions, and accessories. However, many of these datasets have been retracted \cite{boutros2023synthetic} (e.g., VGGFace \cite{Parkhi15vggface}, MS1M \cite{guo2016ms1}, MegaFace\cite{kemelmacher2016megaface})  rendering the remaining datasets scarce. 
The datasets still available are limited by several factors.\\
\textbf{Privacy and Ethical Concerns}: The collection and use of facial images raise numerous privacy and ethical issues, which must be carefully addressed to comply with data protection regulations and ensure the responsible use of face recognition technology.\\

\textbf{Data bias}: Real-world datasets often suffer from imbalanced distributions of different demographic attributes or environmental conditions (.e.g, camera orientations, light conditions). This can lead to biased models that perform poorly on underrepresented ethnic, age or gender groups or challenging scenarios\cite{wang2019racial}.\\

\textbf{Annotation Quality}: The accuracy of face recognition systems relies heavily on the quality of the annotations in the training dataset. Essentially, each class must contain only additional images from the same identity. Manual annotation is a time-consuming and labor-intensive process that may introduce errors or biases that can adversely affect the performance of the resulting models.\\

These limitations call for an alternative method of procuring data. In this article, we show that the usage of 3D rendered synthetic faces via the Datagen face generation platform \cite{friedman2023knowing, yudkin2022handsup, shadmi2021using}, can outperform recent GAN methods \cite{DiscoFaceGAN,synface} and produce comparable results to those achieved via 3D synthetic data pipelines \cite{Bae2022Digiface}.

The structure of this paper is as follows: In Section \ref{section:related_work}, we discuss previous related work. In Section \ref{section:methods}, we provide details about the dataset generation and training paradigms for all our experiments. Section \ref{section:experiments} presents our  experiments and the results associated with each experiment. In Section \ref{section:discussion}, we discuss the results and their broader implications. Finally, in Section \ref{section:future_work}, we outline potential future work that we believe is necessary in the domain of synthetic-based face recognition to further boost current performance.

The contributions of this paper are as follows: 
\begin{itemize}[leftmargin=10pt]
\item Our model attains results that are on par with the current state-of-the-art, and by leveraging the granular control our platform offers, we demonstrate the significance of intra-class variance. This is achieved by incorporating 3D rendered assets such as hats, makeup, object occlusions, hand occlusions, haircuts, and hair color changes, which contribute to the overall accuracy of our model.
\item We illustrate that by using a limited number of real images and identities, our model can achieve results comparable to those obtained by models trained on hundreds of thousands of real images. Specifically, we obtain an accuracy of 98.7\% on LFW \cite{LFWTech}, whereas the current real-data SOTA is 99.86\% for LFW  (see Table \ref{tab:SOTA comparison}). 

\item We highlight how controlled data generation can contribute to a better understanding of the essential features for effective face-recognition algorithms. Specifically, we provide evidence of the importance of varied eyebrows by subsampling a small number of eyebrows from our dataset and showing that models trained on real data are highly susceptible to eyebrow structure variations. 
\end{itemize}


\begin{table*}
	\centering
	\begin{tabular}{lllllll}
		\toprule
		\cmidrule(r){1-2}
		Data & Model &  Dataset & LFW (\%) & CFP-FP (\%) & Age-DB (\%) & Average (\%)\\
		\midrule
		DigiFace & ArcFace & 10Kx72 & 93.43 & \textbf{85.6} & 74.58 & 84.53\\
		Ours & ArcFace & 29.7Kx20 & \textbf{94.91} & 83.38 & \textbf{77.58} & \textbf{85.29}\\
		\midrule
		DigiFace & AdaFace & 10Kx72+100KX5 & 95.40 & 88.77 & 79.72 & 87.96\\
		\bottomrule
	\end{tabular}
	\caption{Pure synthetic training results. First row illustrate the results of training with DigiFace dataset in our own arcFace pipeline. Second row illustrates training with Datagen data using our pipeline, third row are the highest results reported in \cite{Bae2022Digiface}. Comparing to DigiFace on our flow (trained with arcFace) we achieve superior results of by 1.48\% on LFW and 3\% on AgeDB. Compared to previously submitted results on AdaFace trained on 1.22 million DigiFace images, our results fall behind on 0.49\% and 2.14\% on LFW and AgeDB respectively. Results on CFP-FP are lower both on our pipeline and the published results. This can be attributed to a relatively low amount of extreme yaw images.}
	\label{tab:Pure Synthetic data}
\end{table*}

\begin{table*}
	\centering
	\begin{tabular}{lllllll}
		\toprule
		\cmidrule(r){1-2}
		Method & Model & Dataset & Real Images & LFW (\%) & CFP-FP (\%) & Age-DB (\%)\\
		\midrule
	     Ours 600K & ArcFace & 29.7KX20 & 40K (2Kx20) & 98.37 & 90.93 & 88.98 \\
        DigiFace 500K & AdaFace & 10Kx50 & 40K (2Kx20) & 99.05 & 94.01 & 89.77\\
        \bottomrule
        DigiFace 1.22M & AdaFace & (10Kx72 + 100Kx5) & 40K (2Kx20) & 99.17 & 94.63 & 90.50 \\
        \bottomrule

	\end{tabular}
	\caption{Comparison to DigiFace synthetic SOTA after finetuning. Results show that following finetuning with the same amount of data, we achieve close to SOTA results on LFW and CFP-FP datasets, falling behind on 0.68\% and 0.79\% respectively. These margins might be attributed to the different models (Arcface in ours vs Adaface) used for the training, as adaface showed superior published results on CFP-FP \cite{adaface}.}
	\label{tab:SOTA comparison}
\end{table*}

\begin{figure}[H]
\centering
    \includegraphics[width=\linewidth]{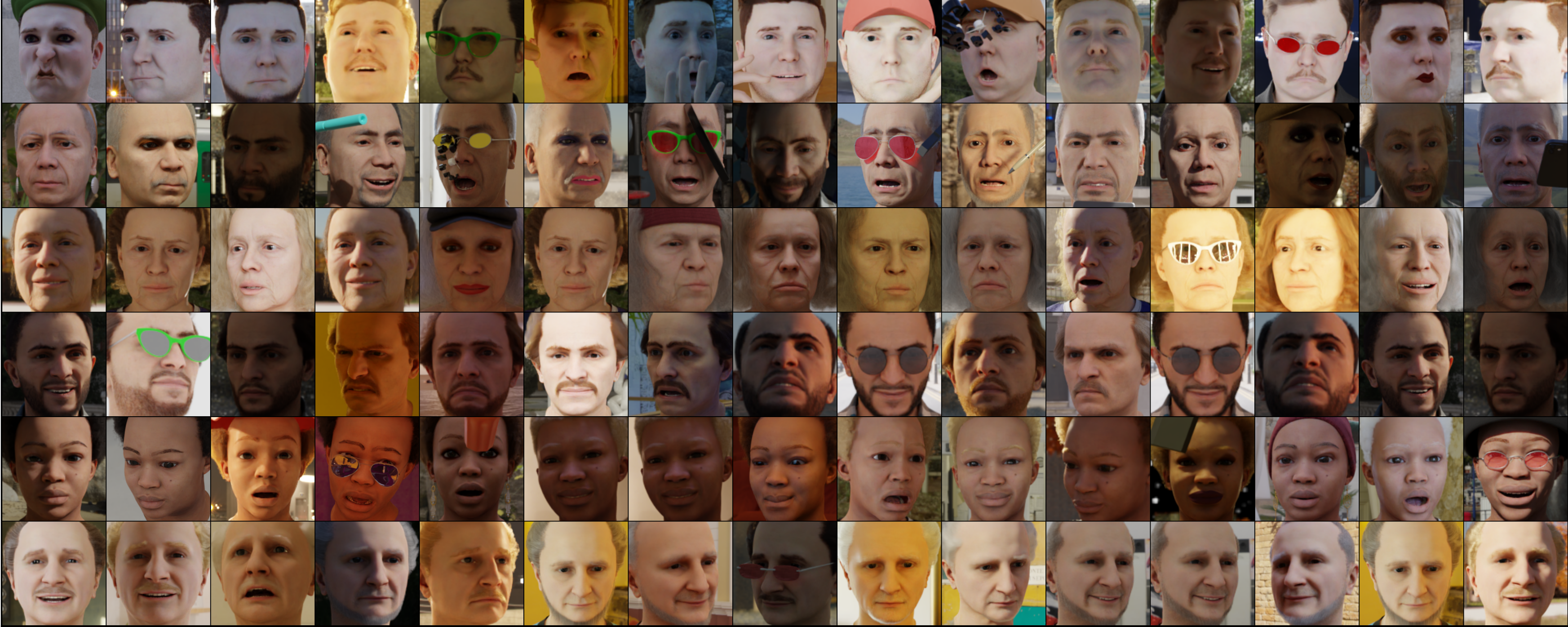}
    \caption{Example of the variability in our dataset for six different identities (a row per identity). Intra-class variance is enhanced by different assets (occlusions, hats, makeup, glasses, facial hair, hair color, hair-cut) as well as the varied poses, background and lighting conditions.}
    \label{fig:general variance}
\end{figure}
\section{Related Work}
\label{section:related_work}
\textbf{Publicly released real faces datasets.} Publicly available datasets satisfy conditions such as pose variability, image quality conditions, lightning conditions, and accessories. Current available datasets include WebFace260M, which comprises 260 million images of 4 million identities \cite{zhu2021webface260m}, IMDbFace that contains 1.7 million images of 59,000 identities \cite{wang2018devilIMDB}, MegaFace2 with 4.7 million images from 672,000 identities \cite{MegaGace}, the CASIA-Webface dataset, which comprised about 500,000 images spanning roughly 10,500 identities, \cite{casiawebface} and the Glint360K dataset, containing a substantial volume of 17 million images across 360,000 identities \cite{wang2022survey, boutros2023synthetic}. 
MS1M, another dataset that originally held approximately 10 million images of 100,000 celebrity identities, was retracted due to a high percentage of noise \cite{guo2016ms1}. MS1MV1 and MS1MV2, the cleansed versions of MS1M, included approximately 3.8 million and 5.8 million images of 85,000 celebrity identities, respectively \cite{deng2017marginalMS1V1, deng2019arcface}. Additional widely used datasets are no longer available such as VGGFace \cite{Parkhi15vggface} and MegaFace\cite{kemelmacher2016megaface}) \cite{boutros2023synthetic} rendering the remaining datasets scarce.

\textbf{Generative models based Face generation.}
A dominant member of the deep generative algorithms, GANs \cite{goodfwllow2014} are used also in the domain of data generation for face recognition training \cite{bao2018facesynth, shen2018faceid, hou2023Deepgenerative, DiscoFaceGAN}.  SynFace\cite{synface} reached an accuracy of 88.98\% on LFW by employing the GAN based model DiscoFaceGAN \cite{DiscoFaceGAN} to generate a training dataset consisting of 10K identities with 50 images per identity. The results were further improved to 91.97\% by applying Identity Mixup (IM) in the form of linear interpolation between two identities in the embedded space, indicating that the learning algorithm can be challenged to better perform with identities that are close in the embedding space. Mixing the dataset with additional 2K real identities further increased the results up to 95.78\%. DiscoFaceGan results were limited by two main factors, the algorithm struggles maintaining 3D consistency \cite{GRAM} on variable poses, and there is a limitation to the model's intra-class variance, most severely in its ability to generate variable facial expressions \cite{synface}. SFace \cite{boutros2022sface} reached an accuracy of 91.87\% on LFW for pure synthetic data. Using additional Knowledge transfer from a model trained on real data reached 98.5\%, while combining both approaches (knowledge transfer and regular classification training) reached 99.13\%. However, this technique requires a pretrained face recognition model (.e.g., FaceNet \cite{schroff2015facenet} and so it is not purely trained on synthetic data).


Diffusion models (DM) \cite{MasterPaperStableDiff, ramesh2022hierarchical, Rombach_2022_CVPR_latentSD} have gained increasing popularity with a fast growing community and visually striking results. As of writing the article, there is a single study comparing the ability of different DM models to create realistic and diverse faces. In the experiment, the author generates data from different models, transforms the images to an ImageNet \cite{deng2009imagenet} embedded space, and calculates the Fréchet Inception Distance (FID) between the embeddings of the generated and real face images. As a baseline, the author splits the 10k real images into two sets and calculates the FID between them. As expected, the comparison between real face images will receive the lowest FID score, as they are from the same distribution. Results show that at the time of writing the article, DM generated face images received much higher FID scores (approximately x5 higher then the baseline score for the best model) indicating that DM generated face images are still not comparable to real images. \cite{borji2022comparisonSD}.

\textbf{3D Rendered Face face generation.}
Microsoft released a synthetic dataset \cite{Bae2022Digiface} consisting of 1.22M images with 110K identities, reaching a final accuracy of 96.17\% on LFW trained on AdaFace\cite{adaface} with a backbone of Resnet 100 \cite{ResNet}. Further more, they showed that using aggressive augmentations can help reduce the gap between real and simulated data, showing an increase from 88.07\% to 94.55\% in accuracy. DigiFace \cite{Bae2022Digiface} was generated using the pipeline introduced at Wood et al. \cite{wood2021fake}, using a generative model learned from 3D 511 unique individuals to generate a total of 110K identities.
Out of the pre-mentioned methods, our data generation platform most resembles that one used in order to create the DigiFace dataset. 
 

\section{Methods}
\label{section:methods}
\subsection{Dataset Generation}
Our dataset was generated using the Datagen \cite{friedman2023knowing, yudkin2022handsup, shadmi2021using} face generation SDK. The platform uses a physically-based rendering engine that renders 2D images from 3D mesh and texture models. The SDK enables easy creation of any desired distribution. Each datapoint consists of the RGB visible spectrum image, with additional meta-data and labels (e.g., key-points, segmentation maps, depth maps, normal maps, and more). 

For this article, we sampled a subset of 30,000 identities from the identity pool (see figure \ref{fig:mosiac}. Our demographics consisted of North European (68.82\%), African (8.52\%), Hispanic (7.94\%), Mediterranean (6.38\%), Southeast Asian (5.01\%), South Asian (3.32\%). 

For each identity, we generated 20 samples of 256x256 or 512x512 resolutions. Both the camera and the human were rotated with yaw, pitch and roll according to approximately normal distributions (compounded of several normal distributions) of mean 0, and variance of of 25\degree, 10\degree, 2.5\degree respectively. HDRI background was sampled randomly among Daytime, evening and night, following by a uniform rotation between [0, 360]\degree. This HDRI rotation process adds two types of variability. First, it changes the perceived background in the generated image. Second, it alters the direction of the light source, which in turn impacts the light and shadows that are reflected on the actor.
For each sample, the expression was randomly sampled from our available presets (i.e., neutral, happiness, sadness, surprise, anger, fear, contempt, disgust and mouth open). All expressions with equal probability to appear. Every identity in our platform was associated with a specific default eye color, iris shape (texture), and eyebrow style during generation. We retained these default values for each identity, as they ensured uniqueness among the different identities in our pool. 

\begin{figure*}[t]
    \centering
    \includegraphics[width=1\textwidth]{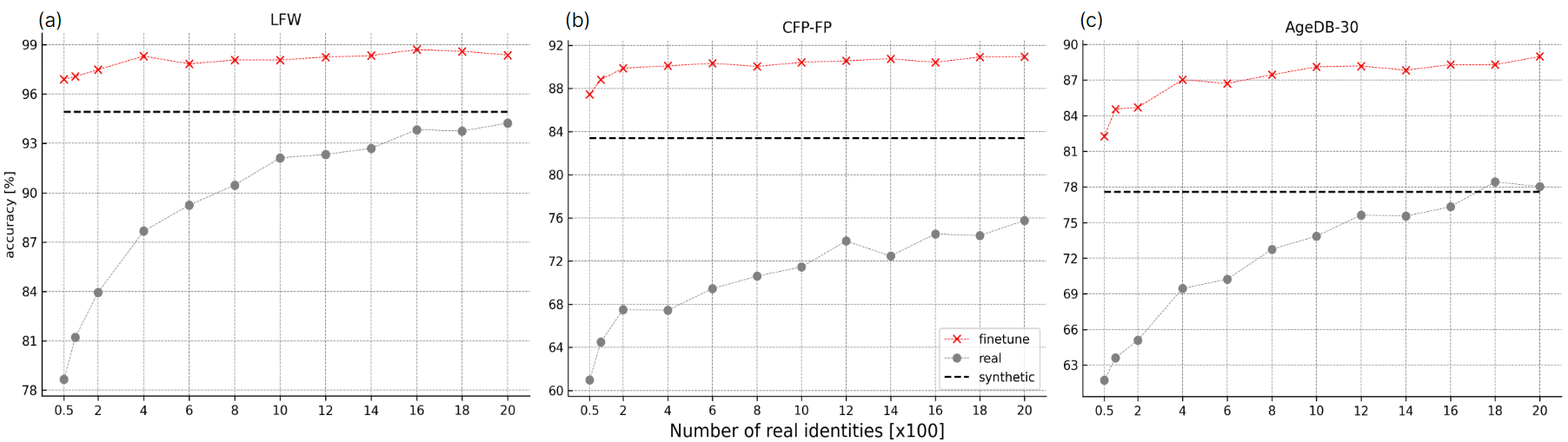}
    \caption{Finetuned vs. real results on (a) LFW, (b) CFP-FP, and (c) AgeDB datasets. The synthetic model was trained on 29K identities, with 20 samples per ID (dashed line). Fine-tuned models (red dots) were with a varying number of identities (represented by the x-axis). For reference, we also trained a model with each of the real sample batches (gray dots). Pure synthetic model outperforms training on batches of real data within the examined range (up to 20K samples). In addition, fine-tuning on allows significant improvement in results relative to pure synthetic training even with a very small amount of real data.} 
    \label{fig:finetune}
\end{figure*}


\begin{table*}[t]
	\centering
	\begin{tabular}{llllll}
		\toprule
		\cmidrule(r){1-2}
		Experiment & Dataset & LFW (\%) & CFP-FP (\%) & Age-DB (\%) & Average (\%)\\
		\midrule
	     Hair Variability & Baseline & 94.5 & \textbf{83.87} & 74.67 & 84.34\\
		                  & Hair-Cut Variability  & \textbf{94.91} & 83.38 & \textbf{77.58} & \textbf{85.29}\\
		\bottomrule

	\end{tabular}
    \caption{Incorporating Additional Hair Variability. The addition of hair variability resulted in an improvement in the LFW and, most notably, the Age-DB metrics. The significant increase in accuracy for Age-DB could be attributed to the high variability inherent in this database, as hairstyles tend to undergo considerable changes over an individual's lifespan.}
	\label{tab:hair variability results}
\end{table*}

Each male sample was generated with 15\% probability to receive a beard. Glasses were samples with 15\% chance of appearing, regardless of gender. 
Eye gaze direction was also adjusted, uniformly sampled with horizontal sides between [-0.5, 0.5] and vertical sides between [0.85, 1] meters. The gaze distance was also sampled, ranging between [0.3, 6] meters.

Hair color for each sample was also modified, relative to the identity's default values for melanin, whiteness, roughness, and redness, with uniform changes within a range of $\pm25$\%.

Additional variability was generated by randomly adding makeup, occlusions, hats and randomized expressions. (see Figure \ref{fig:general variance}) These additions were used for creating two different batches of data. First batch contained a single addition from the above list with probabilities of 3\%, 2.5\%, and 3.5\%, for makeup, occlusions and hats, respectively. In the second batch which constitutes 17\% of the data we allowed simultaneous additions of makeup, occlusions, hats, and additional randomized expressions, each generated with a probability of 15\%, 50\%, 70\%, and 50\%, respectively.\footnote{For more information about our platform see \url{https://datagen.tech}} The randomized expressions were added in order to increase variance on-top of our platform presets and were defined by randomly sampling a single or two action units (one for the eyes, and one for the mouth), with identical probabilities.



\subsection{Training}
All models in this study were trained using the ArcFace loss \cite{deng2019arcface}, incorporating a margin of 0.5 and a scale of 64 with an IResNet50 architecture  \cite{ResNet} backbone. Models were trained on a single 16GB NVIDIA Tesla-4 GPU with batch size set to 256 for 24 epochs with a multi-step learning rate decay by a factor of 0.1 at milestones 10, 18, and 22. In order to be as similar to our validation and test data prepossessing, we utilized the RetinaFace detector \cite{deng2020retinaface} for facial bounding box extraction, as opposed to using the facial bounding box modality provided by the Datagen platform. The key-point modalities were then applied to perform face alignment using the similarity transform (scale, rotation and translation). 
Images were resized to 112x112 and normalized, with a mean of 0 and standard deviation of 0.5 for all channels. All the code was implemented using pytorch\cite{pytorch_NEURIPS2019}.

\textbf{Evaluation Protocol.} Our study employs the open-set protocol for evaluating the model's performance \cite{sphereface}. This approach entails using disjoint identities for testing, ensuring that they are not present in the training set. Our primary aim is to address the problem of face verification, which involves comparing pairs of facial images to ascertain if they originate from the same individual. During the test phase, we apply 10-fold cross-validation on our test set, deriving the threshold from the 9 folds and applying it to the remaining fold. The face verification average accuracy is reported on LFW\cite{LFWTech}, CFP-FP\cite{cfp-paper} and AgeDB\cite{moschoglou2017agedb} benchmark datasets.

\textbf{Data Augmentations.} 
Augmentations were adapted from \cite{Bae2022Digiface} and implemented via the albumentations python package \cite{albumentations}. More specifically, we used horizontal flip with a probability of 0.5 (p=0.5),  conversion to gray scale (p=0.1), Gaussian blur (p=0.05), Gaussian noise (p=0.035), motion blur (p=0.05), JPEG compression (p=0.05), downscale and upscale (p=0.01) and color jitter (p=0.1) with brightness within [0,0.15], contrast within [0,0.3] and hue within [0,0.1] saturation [0,0.1] (where all ranges indicate uniform sampling).

\textbf{Fine-tuning.} 
When finetuning our model, we used our pre-trained backbone and replaced the arcface head to consist of the fine-tune number of parameters. Learning rates were adjusted as in DigiFace \cite{Bae2022Digiface} so that the backbone will train with lr/100 and the head with lr/10. The learning schedule and number of epochs remained the same as the regular training regime. 

\begin{table*}
	\centering
	\begin{tabular}{llllll}
		\toprule
		\cmidrule(r){1-2}
		Experiment & Dataset & LFW (\%) & CFP-FP (\%) & Age-DB (\%) & Average (\%)\\
		\midrule
	                 & Baseline & 93.15 & 81.91 & 74.08 & 83.04\\
		 Additional variance & Combined  & 93.65 & 82.65 & 74.44 & 83.58\\
		             & Separate & \textbf{94.27} & \textbf{83.8} & \textbf{74.87} & \textbf{84.31} \\
		\bottomrule
	\end{tabular}
	\caption{Additional variance experiment. In order to understand the impact of additional variance either generated together (multiple per image) or separate (single per image) on our model, we generated hats, makeup and occlusions with the probabilities of 3\%, 2.5\% and 3.5\% respectively. The results show that the additional variance, although amassing to only 9\% of our data improved results. Combined generated data, increased results by a lower value although amounting to a total variance of 17\% of the data. This results might indicate that the additional multiple-per-image variance photos provided harder samples for the model to train probably due to multiple occlusions, which the model was not able to generalize well.}
	\label{tab:general variability results}
\end{table*}

\section{Experiments and Results}
\label{section:experiments}
In this section, we present our experimental design and results. 
First, we show our result compared to the current synthetic SOTA. Secondly we present our finetuning results, and debate how they compare to both synthetic and real SOTA. Following that, we show how different intra-class-variance factors are affecting our results, and lastly we show a use-case of how controlled data can be utilized in order to understand the the importance of different face parts in face-recognition systems.

\subsection{Pure synthetic training results}
Our pure synthetic training results are summarized in Table \ref{tab:Pure Synthetic data}. We compare our results to those reported by DigiFace\cite{Bae2022Digiface} which are the current SOTA when training on pure synthetic data to the best of our knowledge, and also compare our results with those obtained by training on the Digiface dataset with our own pipeline.
Compared to previously submitted results on AdaFace trained on 1.22 million DigiFace images, we show comparable results achieving 94.91\% on LFW, 83.38\% on CFP-FP and 77.58\% on AgeDB (rows 2 and 3 in Table \ref{tab:Pure Synthetic data}). Our results fall behind on 0.49\% and 2.14\% on LFW and CFP-FP respectively, this might be attributed to both the different models used, and the different amounts of data. 
When using DigiFace dataset (DigiFace\cite{Bae2022Digiface}) trained on our arcFace pipeline, (rows 1 and 2 in Table \ref{tab:Pure Synthetic data}) we surpass DigiFace by 1.48\% on LFW and 3\% on AgeDB. Results on CFP-FP are lower both on our pipline and the DigiFace published results. This might be attributed to the chosen distribution of yaw in our dataset, that does not include many profile images. The results reported contain all the variance discussed in the method section \ref{section:methods}.

\subsection{Finetune}
As previously mentioned, obtaining a large quantity of real data can be challenging. However, there are situations where a limited number of samples are accessible. To examine the effects of merging real data with our synthetic dataset, we finetune a model that encompasses our full range of variability. To assess the influence of small quantities of real data, we randomly sampled varying number of identities, ranging from 10 to 2000, with 20 samples per identity. The results are summarized in Figure \ref{fig:finetune}. We demonstrate that our model can achieve high accuracy comparable to those trained on hundreds of thousands of real images, even with an extremely small number of real samples. Furthermore, we show that a fine-tuned model's performance significantly exceeds that of a model trained solely on the same amount of real data. A question raised here is whether the increase in accuracy is attributable to the photo-realism gap or to the variance gap (consisting of intra-class and inter-class variability). Assuming that real world variance cannot be encompassed within a thousand images (50 identities with 20 sample per identity) we can observe that a photo-realism gap for face recognition exists, and attributes to a reduction in the error rate by 39.1\% for LFW , 24.5\% for CFP-FP and 30\% for Age-DB as accuracy increases from 94.91\% to 96.9\%, 83.38\% to 87.46\% and 77.58\% to 82.28\% respectively \ref{fig:finetune}. These results are slightly higher then those previously demonstrated on segmentation benchmarks \cite{friedman2023knowing}, and might be attributed to the higher dependencies of face-recognition models on actual rgb pixel values as opposed to relationships between neighboring pixels.  
In Table \ref{tab:SOTA comparison} we compare the results of fine-tuning with 40K real samples to those reported by \cite{Bae2022Digiface}. We achieve competitive results on LFW and CFP-FP falling behind on 0.68\% and 0.79\%, respectively.

\subsection{Effects of generated variance} 
In order to explore the effects of the additional generated variance on our model we conducted two experiments. The first experiment focused on the contribution of hats, occlusions, makeup and randomized additional expressions as intra-class-variance. The second experiment was focused specifically on hair-cut variability. We separated these experiments since we hypothesized that hair-cut variance may have specific contribution to age-DB as it allows simulating significant changes of hair-cut over time often occurring along lifetime. 
For each experiment, we have a baseline and the modified version. In our changed version, we swap baseline images with the samples containing the additional variance (.e.g., an identity had 20 baseline samples, after the swap, it has 15 old samples, and 5 new samples containing a hat). As a result, all samples are the same, except for the swapped samples. In our general variability test (Table \ref{tab:general variability results}) we are examining to see the effects of hats, makeup, occlusions and expressions either combined together (with a high probability of appearing together) or separated (one per image). Our datset consists of 27K unique identities with 20 samples per identity. The results show an increase between the baseline and the combined variability, increasing the averaged accuracy from 83.04\% to 83.58\% (LFW 93.15\% to 93.65\%). Introducing the variability separately increased the results further from 83.04\% to 84.31\% (LFW 93.15\% to 94.27\%).

\begin{figure}[H]
\centering
\includegraphics[width=\linewidth]{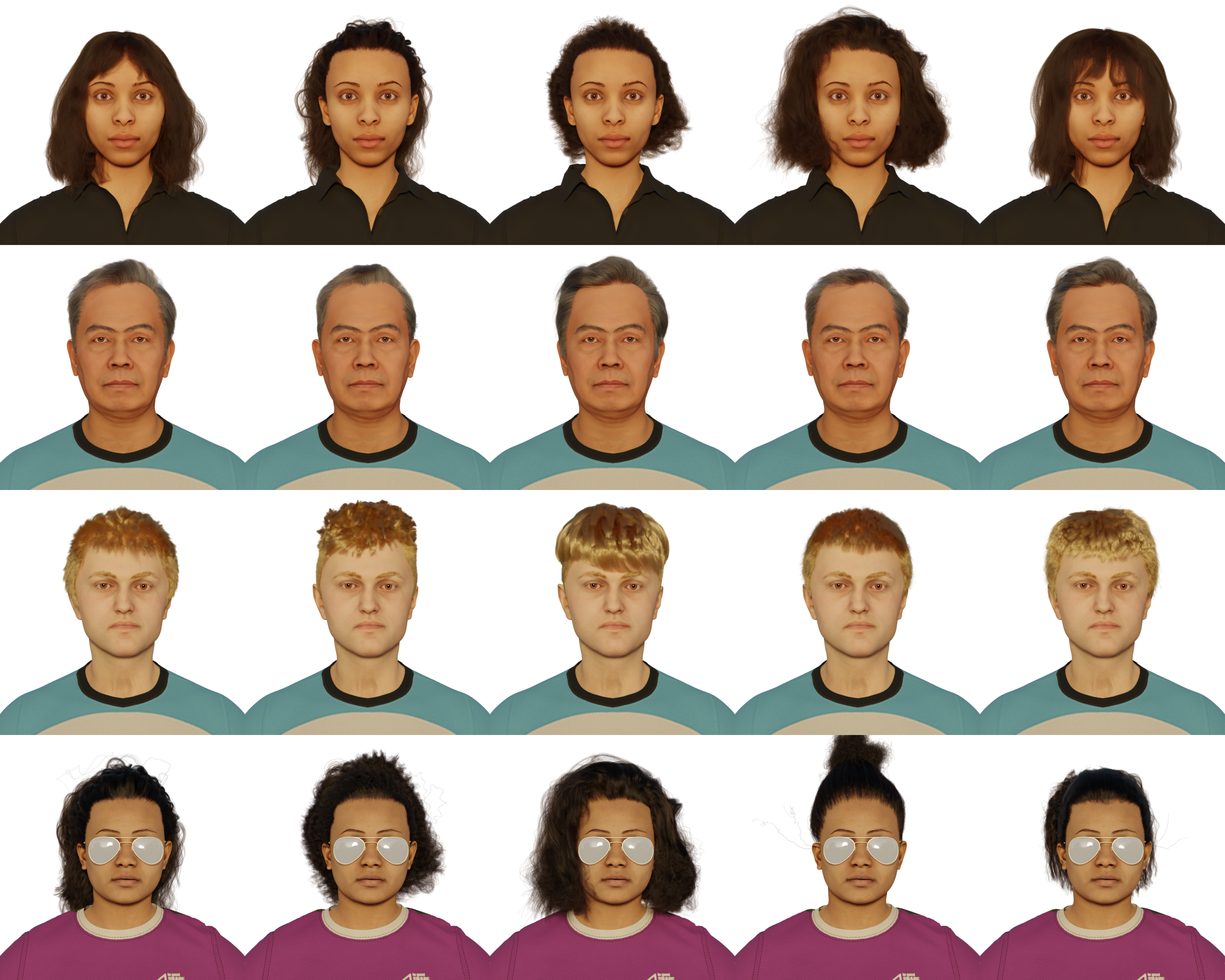}
\caption{Illustration of the different hair style clusters used in our hair variability experiment (each row represents a different cluster). Hair assets were clustered into groups where each group had the same hairline, hair type (e.g., curly, straight, wavy) , thickness (.e.g., fine, medium) and general appearance (e.g., oily, dry).}
\label{fig:hairstyles}
\end{figure}

In the second experiment, we examine the effect of adding hair cut variability to our dataset. Our dataset consists of 29K identities with 20 samples per ID. All the hair assets existing in the platform were clustered into groups of different types. Each group maintained the same hairline, hair type (e.g., curly, straight, wavy), thickness (e.g., fine hair, medium hair, coarse hair), and general appearance (e.g., oily, dry, thin, thick), with the only varying aspect being the haircut itself (see Figure \ref{fig:hairstyles}). As in the previous experiment, there were two datasets, a baseline and the altered dataset, where the altered dataset consisted of the same ids and the majority of the previous samples, with only the samples consisting of the varying hair styles swapped. A total of 32.8\% of the samples where swapped, averaging at 6 samples per identity containing variations of hair-cut. 


The results are summarized in Table \ref{tab:hair variability results} and show that the average accuracy has increased from 84.34\% to 85.29\% (LFW from 94.5\% to 94.91\%). Most notably, the Age-DB accuracy increased by 2.91\%. Age-DB is a diverse test set, featuring images of people at different stages of their lives. This improved performance likely reflects the model's enhanced ability to recognize faces with different hairstyles across multiple life stages, ultimately contributing to the increased accuracy on the Age-DB test set.

\begin{figure*}[t]
    \centering
    \includegraphics[width=1\textwidth]{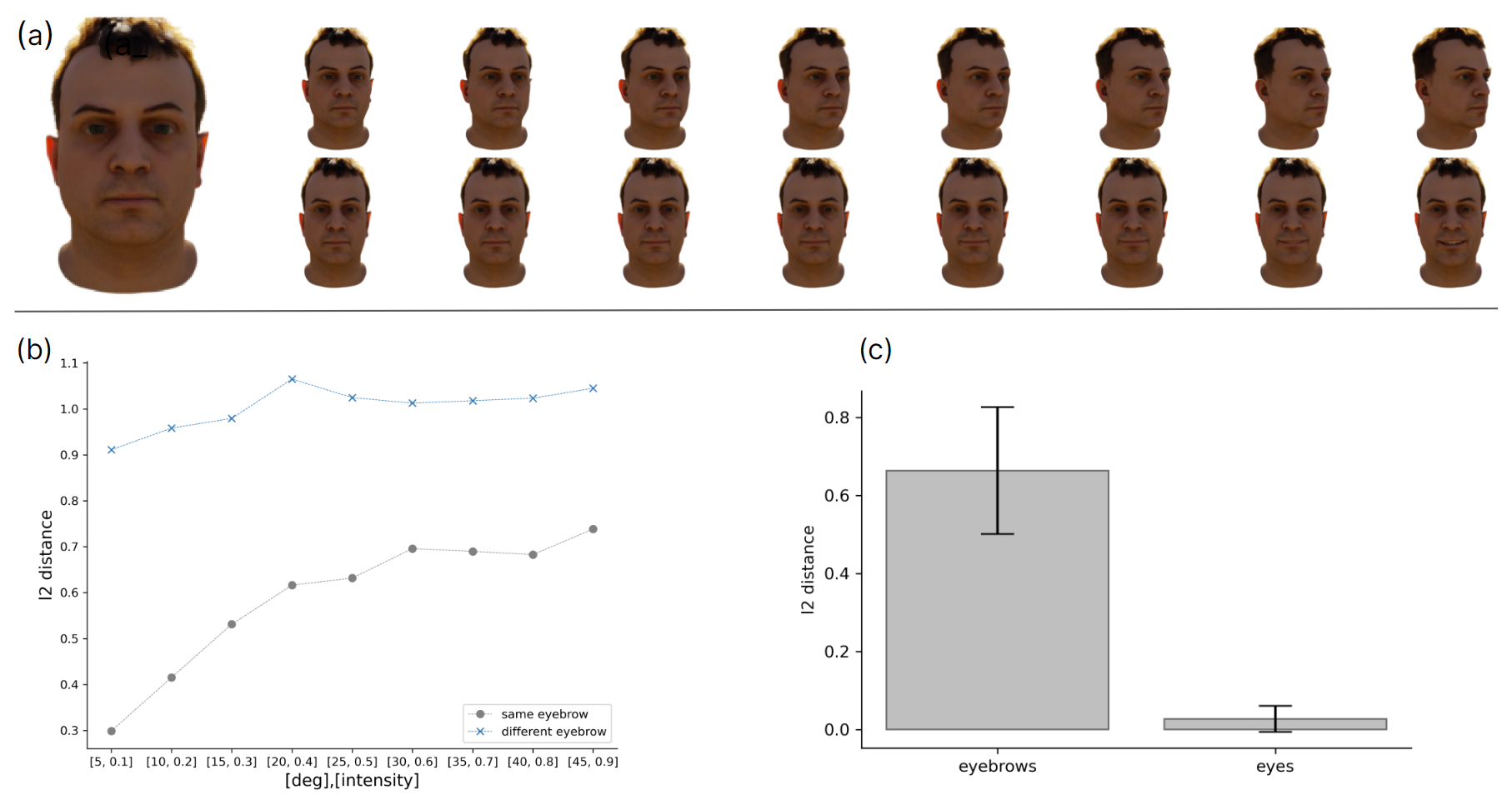}
    \caption{ Alternating face parts. (a) The leftmost image is a simple front-facing enrollment image. The first upper row shows gradual changes in head rotation from 0° to 45°, while the second row illustrates gradual changes in facial expression intensity.  (b) L2 distance sensitivity. Distances between a front facing neutral expression reference and our alternating conditions of yaw and expression intensity. The value of the gray curve are the baseline values considered valid within the intra-class variations while the blue  values, well above this baseline, are prone to change the networks prediction. (c) Averaged l2 distances between a frontal facing, neutral expression with changing eyebrows (left) and eyes (right). A change to the eyebrows results in an average difference of 0.664, while a change to the eyes results in an average difference of 0.027. Overall, we observe that eyebrows are important in the context of facial recognition and alternating their appearance and shape may lead to predicting the photo as another identity, whereas the change in eyes-color and iris textures have much lower influence on the l2 distances probably due to the small face crop sizes used for most face verification models.}
    \label{fig:eyebrows}
\end{figure*}
\subsection{Controlled data use-case}

To underscore the efficacy of using controlled synthetic data, we follow a case study in the context sensitivity to different face parts \cite{facepartsContribution}. This area is important as it increases our understanding of what different facial parts (.e.g., eyes, eyebrows, mouth, etc.) are imperative for a model to accurately classify identities.

For that purpose, we alternated two factors: eyes (colors and iris texture) and eyebrows. 

First, to show how the model reacts to valid intra-class variations (Figure \ref{fig:eyebrows} (b) the grey line). We measure the l2 distance in the  FaceNet\cite{Schroff2015} embedding space between a reference image of frontal pose and neutral expression and a set of varying poses and expressions on the same identity where all other aspects of the image remained similar (background, light conditions etc.,)

Pre-trained models are expected to be agnostic to pose and expression variations and therefore, we refer to this distances as valid intra-class variance, whereas higher l2 distances, occurring due to alternating face parts (eyes and eyebrows) would indicate values that are prone to change the networks prediction (see Figure \ref{fig:eyebrows} a and b).

By retaining the reference, and changing the eyebrow by a single random sampled eyebrow, we see a leap in the l2 distance (see Figure \ref{fig:eyebrows} b blue line). 
The average difference between the two conditions (the gray and blue lines in Figure \ref{fig:eyebrows} b) is 0.415 $\pm 0.107$. Additionally, in the already high intra-class variance cases (e.g., 45\degree and 0.9 intensity), the l2 distance can be above 1. This indicates that that eyebrows appearance is a descriptive factor for face recognition models and alternating it may lead to false rejection if the change is natural as part of styling or true rejection in case of fraud.


In the last experiment, we compare the effect of alternating eyebrows to alternating eyes. We use only frontal facing and neutral expression images in-order to check for the effects of 25 eyebrows and 100 eyes sampled from the platform pool. Different eye samples have different color and iris textures.
We observe that changing the eyebrows alone account for an average increase of 0.664 in l2 distance $\pm 0.16$. In contrast, the model is not sensitive to the eyes, this is expected, as the image sizes in modern face recognition models are usually 112x112 (160x160 for facenet), where the eyes inhabit a small number pixels.


\section{Discussion} 
\label{section:discussion}
In this work, we have demonstrated the potential of using synthetic data for face recognition, particularly by leveraging the controlled environment offered by our 3D rendering pipeline. Our results reveal that our model, trained on synthetic data, can achieve results competitive with the state-of-the-art on multiple benchmark datasets. Moreover, we have shown that incorporating various forms of intra-class-variance in the dataset, such as hairstyles, makeup, hats, and occlusions, can improve the model's performance. This emphasizes the importance of intra-class variance in developing more robust and accurate face recognition models.

We also highlighted the advantage of fine-tuning our model with a limited number of real images. Our experiments indicate that even a small amount of real data can considerably improve the model's performance, achieving results comparable to those obtained by models trained on large-scale real datasets. This finding suggests that our approach can be beneficial in scenarios where obtaining substantial volumes of real data is challenging.

Lastly, we demonstrated the value of controlled data generation in better understanding the essential features of face recognition. By incorporating the separation of variables, we are able to understand our model's weaknesses as well as what is needed to improve recognition by employing synthetic data. Our experiments showed that models trained on real data are highly sensitive to variations in eyebrow structure while not sensitive to eyes color and textures, suggesting that eyebrows can be an important factor in determining identity. This insight can help researchers and practitioners develop more robust and accurate face recognition systems by focusing on such discriminative features.

\section{Future Work}
\label{section:future_work}
While our study has shown promising results, several avenues for future work can be explored to further enhance the efficacy of synthetic data in the face recognition domain. With the growing power of DM models \cite{MasterPaperStableDiff} grows the power of reducing the domain gap, and adding additional variance to controlled 3D synthetic data. Emerging research venues such as image to image text guided translation and impainting \cite{brooks2023instructpix2pix, orgad2023editing, Rombach_2022_CVPR_latentSD} as well as controlled data generation \cite{controlnet} might be utilized to increase the effectiveness of 3D generated data. However, in order for these models to be effective for face recognition tasks, there must be a viable and fast way for unique identity generation and preservation. The area of personalized SD \cite{personalizedHan, e4t} is still in its initial stages and further research is needed for investigating the combination of rendered data and diffusion models. 
An additional significant gap in deep face recognition pertains to aging \cite{sawant2019age,aging2022}. The challenge arises from the natural biological transformations that occur throughout our lifetimes. These alterations, which influence the overall facial structure, including changes in the jawline, ears, nose shape, addition of wrinkles and age spots and more, complicate the task of maintaining consistent and accurate recognition.
There is a growing need in generating synthetic data with reliable aging simulation.

\bibliographystyle{unsrtnat}
\bibliography{references}
\end{multicols}

\appendix
\balance
\end{document}